\documentclass[journal]{IEEEtran}

\usepackage[english]{babel}
\usepackage{amsmath}
\usepackage{amssymb}
\usepackage[all]{xy}
\usepackage{comment}
\usepackage{float}
\usepackage{graphicx}
\usepackage{subfigure}
\usepackage{float}
\usepackage{xspace}
\usepackage{tikz}

\usepackage{epstopdf}

\usetikzlibrary{trees}

\newenvironment{definition}[1][Definition]{\begin{trivlist}
\item[\hskip \labelsep {\bfseries #1}]}{\end{trivlist}}

\newcommand{\qed}{\nobreak \ifvmode \relax \else
      \ifdim\lastskip<1.5em \hskip-\lastskip
      \hskip1.5em plus0em minus0.5em \fi \nobreak
      \vrule height0.75em width0.5em depth0.25em\fi}

\title{An Efficient Genetic Programming System with Geometric Semantic Operators
and its Application to Human Oral Bioavailability Prediction}

\author{\IEEEauthorblockN{Mauro Castelli\IEEEauthorrefmark{1}, Luca Manzoni\IEEEauthorrefmark{3},
Leonardo Vanneschi \IEEEauthorrefmark{2}\IEEEauthorrefmark{1}\IEEEauthorrefmark{3}}\\
\IEEEauthorblockA{\IEEEauthorrefmark{1} INESC-ID, 1000-029 Lisboa, Portugal\\
\IEEEauthorrefmark{2} ISEGI, Universidade Nova de Lisboa, 1070-312 Lisboa, Portugal\\
\IEEEauthorrefmark{3} DISCo, University of Milano-Bicocca, 20126 Milano, Italy\\
 Email: \{castelli,luca.manzoni,vanneschi\}@disco.unimib.it
}}
\date{\today}

\begin{document}

\maketitle

\begin{abstract}
    Very recently new genetic operators, called geometric semantic operators, have been defined for genetic programming.  Contrarily to standard genetic operators, which are uniquely based on the syntax of the individuals, these new operators are based on their semantics, meaning with it the set of input-output pairs on training data.  Furthermore, these operators present the interesting property of inducing a unimodal fitness landscape for every problem that consists in finding a match between given input and output data (for instance regression and classification).  Nevertheless, the current definition of these operators has a serious limitation: they impose an exponential growth in the size of the individuals in the population, so their use is impossible in practice.  This paper is intended to overcome this limitation, presenting a new genetic programming system that implements geometric semantic operators in an extremely efficient way.  To demonstrate the power of the proposed system, we use it to solve a complex real-life application in the field of pharmacokinetic: the prediction of the human oral bioavailability of potential new drugs.  Besides the excellent performances on training data, which were expected because the fitness landscape is unimodal, we also report an excellent generalization ability of the proposed system, at least for the studied application.  In fact, it outperforms standard genetic programming and a wide set of other well-known machine learning methods.
\end{abstract}


%%%%%%%%%%%%%%%%
% Introduction %
%%%%%%%%%%%%%%%%
\section{Introduction}
\label{intro} Genetic Programming (GP)~\cite{poli08:fieldguide,koza:book} is the youngest paradigm inside the computational intelligence research area called Evolutionary Computation (EC) and consists in the automated learning of computer programs by means of a process mimicking Darwinian evolution.  A GP algorithm works by maintaining and evolving a set (often called {\it population}) of so called {\it individuals}, each of which representing a program that is a potential solution to a problem.  A {\it fitness} function (sometimes also called cost or quality function), defined over the space of all individuals, quantifies the ability of each one of them in solving the problem.  After a (typically) random initialisation of the population, the evolutionary process, aimed at progressively improving the fitness of the individuals in the population, takes place by iterating two phases: selection (where the most promising solutions are probabilistically chosen for mating), and the application of the genetic operators (used to explore the space of solutions), which typically consist in crossover (that exchanges parts of two parent solutions in order to generate new offspring) and mutation (that randomly modifies parts of some solutions).

In the last few years, GP has been extensively used both in Industry and Academia and it has produced a wide set of results that have been defined {\it human-competitive}~\cite{Koza:1999:GPI:553446}.  These results cover a wide variety of applicative domains, including quantum computing circuits, analog electrical circuits, design of antennas, mechanical systems, photonic systems, optical lens systems and sorting networks.

While these results have demonstrated the suitability of GP in tackling real-life problems, research has recently focused on developing new variants of GP in order to further improve its performances. In particular, efforts have been dedicated to an aspect that was only marginally considered up to some years ago: the definition of methods based on the semantics of the solutions ~\cite{Uy:2009:ISICA,McPhee:2008,Bryant:1986:GAB:6432.6433,Beadle:2008:CEC,Quang:2011:GPEM,DBLP:conf-gecco-KrawiecL09,Beadle:2009:cec,Beadle:2009:GPEM,Jackson:2010:EuroGP,Jackson:2010:PPSN}.  Though there is no universally accepted definition of semantics in GP, this term often refers to the behavior of a program, once it is executed on a set of data.  For this reason, in many references, the term semantics is intended as the set of input-output pairs on the training data, and this is the terminology that we will adopt in this paper as well.

In this research track, very recently, new genetic operators, called geometric semantic genetic operators, have been proposed in~\cite{PaperMoraglio}.  These operators have the interesting property of inducing a unimodal fitness landscape on any problem consisting in finding the match between a set of input data and a set of known output ones. Classification and regression (which are typical applications of Evolutionary Computing in general~\cite{f2,f7} and GP in particular~\cite{f5,f6}) are examples of this kind of problem. According to the theory of fitness landscapes~\cite{langdon:fogp} (which are briefly introduced later in this paper), this should allow GP to easily solve all these problem.

Nevertheless, as stated in~\cite{PaperMoraglio}, these operators also have a serious limitation: they construct offspring that are bigger than their parents, and this makes the size of the individuals in the population grow exponentially with generations.  In this way, after few generations, the population is composed by individuals that are so big that their fitness evaluation is unmanageable.  This limitation makes these operators impossible to use in practice.

The solution suggested in~\cite{PaperMoraglio} as a future work is to integrate in the GP algorithm a "simplification" phase, aimed at transforming each individual in the population into an equivalent (i.e. with the same semantics), but smaller program.  However, also this solution has some problems: according to the language used to code individuals, simplification can be very difficult, and it is often a very time consuming task.

For this reason, in this paper we propose a completely different solution to this problem, by presenting a new GP system incorporating an implementation of geometric semantic genetic operators that not only makes them usable in practice, but even very efficient, without requiring any simplification of the individuals during the GP run.  In this way, we are for the first time able to exploit the great potentialities of these operators, consisting in the fact that they induce unimodal fitness landscapes.

In order to experimentally validate our new GP system, we have applied it to a complex real-life problem in the field of pharmacokinetic: the prediction of human oral bioavailability of new potential drugs.  The results we have obtained have been compared not only to the ones returned by standard GP, but also to the ones of several other state of the art Machine Learning methods reported in~\cite{springerlink:10.1007/s10710-007-9040-z}.

The paper is organized as follows: Section~\ref{sec:sem-op} presents the state of the art concerning the use of semantics attempting to improve GP.  Section~\ref{sec:land} introduces the concept of fitness landscape, that will be useful successively to explain the potentialities of geometric semantic operators.  Section~\ref{sec:operatoriMoraglio} describes the geometric semantic operators, shows that the fitness landscapes induced by them are unimodal and outlines their limitations.  Section~\ref{sec:implementazione} presents our new GP system that overcomes the current limitations of geometric semantic operators, making them usable (and efficient) for real-life applications.  Section~\ref{sec:results} presents the test problem used, the experimental settings and the obtained results.  Finally, Section~\ref{sec:conclusioni} concludes the paper and suggests some hints for future research.

%%%%%%%%%%%%%%%%%%%%%%%%%%%%%%%%%%%%%%%%
% Semantic Operators: State of the Art %
%%%%%%%%%%%%%%%%%%%%%%%%%%%%%%%%%%%%%%%%
\section{State of the art on the use of Semantics in GP}
\label{sec:sem-op}

The genetic operators in GP systems are usually designed with the only constraint of ensuring syntactic closure, i.e. producing syntactically valid offspring from any syntactically valid parent(s). As stated in~\cite{Uy:2009:ISICA}, using such purely syntactical genetic operators, GP evolutionary search is conducted on the syntactical space of programs, with the only semantic guidance offered by the fitness function employed by selection.

As stated before, GP has been used to successfully solve real life problems; however the usage of purely syntactical genetic operators it is not able to describe the entire dynamic of the evolutionary process. Thus incorporating semantic awareness in the GP process could improve performance, extending its applicability to problems that are difficult with purely syntactic approaches.  Under this perspective, several recent works have proposed the definition of semantic based methods.

These methods appeared in combination with crossover. McPhee et al.~\cite{McPhee:2008} used truth tables to analyze behavioral changes in crossover for boolean problems. They considered the semantics of two components in each tree: semantics of subtrees and semantics of context (the remainder of an individual after removing a subtree). They experimentally measured the variation of these semantic components throughout the GP evolutionary process. They payed special attention to fixed-semantic subtrees: subtrees such that the semantics of the entire tree does not change when they are replaced by another subtree. They showed that there may be many such fixed semantic subtrees when the tree size increases during evolution; thus it becomes very difficult to change the semantics of trees with crossover and mutation once the trees have become large.

While it is possible to represent behavior using truth tables, a more efficient technique is that of using reduced ordered binary decision diagrams (ROBDDs)~\cite{Bryant:1986:GAB:6432.6433} to create reduced canonical representations to measure behavioral difference.

In~\cite{Beadle:2008:CEC} semantic is used to define an algorithm called Semantically Driven Crossover (SDC). The SDC algorithm has been developed based on analysis of the behavioral changes caused by crossover. The key feature of this method is the use of a canonical representation of members of the population (reduced ordered binary decision diagrams-ROBDDs) to check for semantic equivalence without having to access the fitness function. Two trees are semantically equivalent if and only if they reduce to the same ROBDD. This is used to determine which participating individuals are copied into the next generation. If the offspring are semantically equivalent to their parents, the children are discarded and the crossover is repeated. This process is iterated until semantically different children are found. The authors argue that this results in increased semantic diversity in the evolving population, and a consequent improvement in the GP performance.

In~\cite{mcphee:2007:wps32} a new mechanism for studying the impact of subtree crossover in terms of semantic building blocks is proposed. This approach allows to completely and compactly describe the semantic action of crossover, and provides insight into what does (or does not) make crossover effective. Results make it clear that a very high proportion of crossover events (typically over 75\% in the presented experiments) are guaranteed to perform no immediately useful search in the semantic space.

In~\cite{q.u.nguyen-etal:ppsn2010} the authors investigate the role of syntactic locality and semantic locality of crossover in GP. The results show that improving syntactic locality reduces code growth, and that leads to a slight improvement of the ability to generalize. By comparison, improving semantic locality significantly enhances GP performance, reduces code growth and substantially improves the ability of GP to generalize.

In~\cite{Nguyen:2009:SAC:1533497.1533524} the authors proposed Semantics Aware Crossover (SAC), a crossover operator promoting semantic diversity, based on checking semantic equivalence of subtrees. It showed limited improvement on some test problems; it was subsequently extended to Semantic Similarity based Crossover (SSC)~\cite{Quang:2011:GPEM}, which turned out to perform better than both standard crossover and SAC~\cite{Quang:2011:GPEM}. In particular, authors aim to incorporate semantics into the design of new crossover operators, so as to maintain greater semantic diversity and provide higher locality than standard crossover. The idea of SSC was then extended to mutation leading to a counterpart semantic mutation: Semantic Similarity based Mutation (SSM)~\cite{Nguyen:2009:MENDEL}. The experimental results in~\cite{Nguyen:2009:MENDEL} confirm the superior performance of SSM compared to standard mutation.

In~\cite{Beadle:2009:cec} semantics is used to test the effects of behavioral control at the point of the mutation operator. Using semantic analysis, authors present a technique known as semantically driven mutation (SDM), which can explicitly detect and apply behavioural changes caused by the syntactic modifications in programs caused by mutation. The SDM algorithm does not allow mutated programs to be produced when they are behaviorally equivalent to the original program. The aim of this is to avoid getting stuck in areas of the search space that have already been investigated. As in~\cite{Beadle:2008:CEC}, the key feature of the semantically driven operator is the ability to canonically represent programs in such a way that it is possible to compare them, looking for equivalent behaviors.

In~\cite{krawiec:2012:EuroGP} the authors proposed a class of crossover operators for genetic programming aimed at making offspring programs semantically intermediate (medial) with respect to parent programs by modifying short fragments of code (subprograms). The approach is applicable to problems that define fitness as a distance between program output and a desired target. Based on that metric, the authors defined two measures of semantic ``mediality'', which they employed to design two crossover operators: one aimed at making the semantic of offspring geometric with respect to the semantic of parents, and the other aimed at making them equidistant to parents' semantics. When compared experimentally with four other crossover operators, both operators lead to success ratio at least as good as for the non-semantic crossovers, and the operator based on equidistance outperformed all others on some test cases.

Krawiec and coworkers in~\cite{DBLP:conf-gecco-KrawiecL09} have used a notion of semantic distance to propose a crossover operator for GP that is approximately a geometric crossover~\cite{moraglio:tio:gecco2004} in the semantic space. In the class of problems considered in~\cite{DBLP:conf-gecco-KrawiecL09}, the fitness function is usually defined as a metric that measures the divergence between target and output values.  As reported in~\cite{DBLP:conf-gecco-KrawiecL09}, metric-based fitness functions are unimodal by definition because such fitness is a distance in the semantic space. Any linear combination of a pair of semantics is guaranteed to be not worse than the worse of them. Authors pointed out that there is no obvious way of exploiting this property due to the complexity of the genotype-phenotype mapping in GP. Thus, the prospects of designing a crossover operator that works in the genotype space and behaves geometrically in the corresponding semantic space are even more gloomy. Hence, rather than guaranteeing the geometric behavior, their operator tries to approximate it by analysing the offspring after it has been bred. This limit is overcome by the geometric semantic operators proposed in~\cite{PaperMoraglio} and described in section~\ref{sec:operatoriMoraglio}. The work in~\cite{PaperMoraglio} introduces a general method to derive exact semantic geometric crossovers and mutations for different problem domains that search directly the semantic space.  However, as already discussed previously in this paper, these operators by construction produce offspring that have approximately the double of the size of their parents (expressed as the total number of tree nodes). As a consequence, the size of the individuals in the population grows exponentially (as proven in~\cite{PaperMoraglio}) and this makes these operators unusable in practice.

%%%%%%%%%%%%%%%%%%%%%%
% Fitness Landscapes %
%%%%%%%%%%%%%%%%%%%%%%
\section{Fitness Landscapes}
\label{sec:land} The concept of fitness landscape was first proposed in~\cite{Wright1932} to study the evolutionary process in Biology. The notion of a fitness landscape underlying the dynamics of evolutionary adaptation optimization has proved to be one of the most powerful concepts in evolutionary theory, and it has been widely used to model the problem difficulty in evolutionary algorithms (EAs)~\cite{Langdon:2004:GAP:1059733.1059740,Vanneschi:thesis}.  As reported in~\cite{springerlink:10.1007/BFb0103571}, implicit in the idea of fitness landscape is a collection of genotypes arranged in an abstract metric space, with each genotype next to those other genotypes which can be reached by a single application of a given genetic operator, as well as the fitness value.

As described, for instance, in~\cite{Nguyen:2011:GECCO}, a fitness landscape can be seen as a three-dimensional map, which may contain peaks and valleys and the problem solver as a short-sighted explorer searching for the highest peak (for maximization problems). They can be formally modeled and are helpful to understand the ability of a searcher like GP to solve a problem. For example, a smooth and regular landscape with a single hill top (i.e.  unimodal) is typical of an easy problem, while the opposite is true for a very rugged (i.e. multimodal) landscape, with many hills which are not as high as the best one. In the latter case, it is more difficult to find solutions (the highest peaks), since the algorithms can be trapped in any local peak.

Even though not without faults, the general knowledge associated to fitness landscapes is the more rugged landscape, the more difficult the problem.  It has been known since Eigen's work~\cite{springerlink:10.1007/BF00623322} that the dynamics of optimization on a landscape depends crucially on detailed structure of the landscape itself. Extensive computer simulations~\cite{Fontana1987123,PhysRevA.40.3301}, have made it very clear that a complete understanding of the dynamics is impossible without a thorough investigation of the underlying landscape~\cite{Eig07}.

In practice, however, the visualization of the whole search space of a problem is difficult, if not even impossible given its generally huge size and the multi-dimensionality of the neighborhoods imposed by canonic genetic operators. Therefore, a number of methods that attempt to describe the relevant features of fitness landscapes by means of numeric indicators have been proposed~\cite{springerlink:10.1007/BF00202749,Vassilev:2000:ICS:1108888.1108891,citeulike:3180341}.  For a complete review on fitness landscapes in EC (and GP in particular) the reader is referred to~\cite{Vanneschi:thesis}.  The most used indicator that relates problem difficulty with the underlying fitness landscape is fitness distance correlation (FDC), studied for GP in~\cite{vanneschi+tomassini:2003:gecco:workshop}.  \emph{FDC} quantifies the difficulty of a problem by expressing the correlation between the fitness of a sample of individuals and their distance to one globally optimal solution.  As we explain in the next section, geometric semantic operators induce unimodal fitness landscapes, characterized by an ideal value of the FDC (i.e. FDC equal to~1), which typically indicates that the problem is easy to solve.  This makes these operators extremely appealing and promising and encouraged us to develop an efficient implementation.

%%%%%%%%%%%%%%%%%%%%%%%%%%%%%%%%
% Geometric Semantic Operators %
%%%%%%%%%%%%%%%%%%%%%%%%%%%%%%%%
\section{Geometric Semantic Operators}
\label{sec:operatoriMoraglio}

While the semantically aware methods cited in Section~\ref{sec:sem-op} often produced superior performances with respect to traditional methods, they are indirect: search operators act on the syntax of the parents to produce offspring, which are successively accepted only if some semantic criterium is satisfied. As reported in~\cite{PaperMoraglio}, this has at least two drawbacks: (i) these implementations are very wasteful as heavily based on trial-and-error; (ii) they do not provide insights on how syntactic and semantic searches relate to each other.

To overcome these drawbacks, in~\cite{PaperMoraglio}, using a formal geometric view on search operators and representations, the authors introduced a novel form of GP that directly searches the space of the underlying semantics of the programs. This perspective provides new insights on the relation between program syntax and semantics, search operators and fitness landscapes, and allows principled formal design of semantic search operators for different classes of problems.

To explain the idea, let us first consider Genetic Algorithms (GAs), which are similar to GP with the major difference that the solutions are fixed length strings of characters and not computer programs. Let us consider a GA problem in which the target solution is known and the fitness of each individual corresponds to its distance to the target (our reasoning holds for any distance measure used). This problem is easy.  In fact, for instance, if we use point mutation, any possible individual different from the global optimum has at least one neighbor (individual resulting from its mutation) that is closer than itself to the target, and thus fitter. So, there are no local optima: the fitness landscape is unimodal.  This is also confirmed by the FDC that is clearly equal to~1, because fitness and distance to the goal are identical.  Similar considerations hold for many types of crossover, including various kinds of geometric crossover~\cite{moraglio:tio:gecco2004}.

Now, let us consider the typical GP problem of finding a function that maps sets of input data into known target ones. As already discussed, regression and classification are particular cases. The fitness of an individual for this problem is typically a distance between its calculated values and the target ones (error measure).  Now, let us assume that we can find a transformation on the syntax of the individuals, whose effect is a random perturbation of one of their calculated values. In other words, let us assume that we are able to transform an individual $G$ into an individual $H$ whose output is like the output of $G$, except for one value, that is randomly perturbed. Under this hypothesis, we are able to map the considered GP problem into the GA problem discussed above. So, this transformation would induce a unimodal fitness landscape with FDC equal to~1 and every problem like the considered one (e.g. regressions and classifications) should be easily solvable by GP. The same also holds for transformations on pairs of solutions that correspond to GAs semantic crossovers.

Under this perspective, the objective of~\cite{PaperMoraglio} was to find operators on the syntactic (or genotypic) space that map well-known operators on the semantic space.  Here we report the definition of geometric semantic operators given in~\cite{PaperMoraglio} for real functions domains, since these are the operators we will use in the experimental phase.  For applications that consider other kinds of data, the reader is referred to~\cite{PaperMoraglio}.

\begin{definition}{\bf (Geometric Semantic Crossover).}
    Given two parent functions $T_1,T_2 : \mathbb{R}^n \to \mathbb{R}$, the geometric semantic crossover returns the real function: $$T_{XO}=(T_1 \cdot T_R) + ((1-T_R)\cdot T_2)$$ where $T_R$ is a random real function whose output values range in the interval $[0,1]$.
\end{definition}

Reference~\cite{PaperMoraglio} formally proves that this operator corresponds to geometric crossover on the semantic space, and thus induces a unimodal fitness landscape.  To constrain $T_R$ in producing values in $[0,1]$ we use the sigmoid function: $T_R = \frac{1}{1+e^{-T_{rand}}}$ where $T_{rand}$ is a random tree with no constraints on the output values.

\begin{definition}{\bf (Geometric Semantic Mutation).}
    Given a parent function $T : \mathbb{R}^n \rightarrow \mathbb{R}$, the geometric semantic mutation with mutation step $ms$ returns the real function:
$$T_M=T+ms\cdot(T_{R1} - T_{R2})$$ where $T_{R1}$ and
$T_{R2}$ are random real functions.
\end{definition}

Reference~\cite{PaperMoraglio} formally proves that this operator corresponds to a box mutation on the semantic space, and induces a unimodal fitness landscape.

We point out that at every step of one of these operators, offspring contain the complete structure of the parents and one or more random trees as subtrees, plus some arithmetic operators: the size of each offspring is thus clearly much larger than the one of their parents.  The exponential growth of the individuals in the population (demonstrated in~\cite{PaperMoraglio}) makes these operators unusable in practice: after a few generations the population becomes unmanageable and the fitness evaluation process becomes unbearably slow.  The solution that is suggested in~\cite{PaperMoraglio} as a future work consists in performing an automatic simplification step after every generation in which the programs are substituted by (hopefully smaller) semantically equivalent ones.  However, this additional step adds to the computational cost of GP and is only a partial solution to the progressive program size growth.  Last but not least, according to the particular language used to code individuals, automatic simplification can be a very hard task.

Due to all these limitations, it is important to make an effort in implementing a framework that will allow an efficient use of the geometric semantic operators.  The objective of this paper is to present a GP implementation that overcomes this limitation, without performing any simplification step and without imposing any particular representation for the individuals (for example the traditional representation of GP individuals as trees can be used).  This implementation is presented in the next section.  For simplicity, from now on, GP using only geometric semantic crossover and mutation to explore the search space will be indicated as GS-GP (Geometric Semantic GP).

%%%%%%%%%%%%%%%%%%%%%
% GP Implementation %
%%%%%%%%%%%%%%%%%%%%%
\section{The Proposed GP Implementation}
\label{sec:implementazione}

The implementation we propose can be described by the following steps:

\begin{itemize}

    \item We create an initial population of (typically random) programs, exactly as in standard GP (let $P$ be the name of this population from now on).

    \item Given that geometric semantic crossover and mutation need the generation of random trees to be used, we create beforehand all the random trees that will be needed during the whole evolution (this pool of random trees will be called $P_{\text{mut}}$ from now on).

    \item During the first generation, at the moment of evaluating the fitness of the individuals, we create two tables that we call $T_{\text{vp}}$ and $T_{\text{vm}}$.  $T_{\text{vp}}$ (respectively $T_{\text{vm}}$) contains, for each individual in $P$ (respectively in $P_{\text{mut}}$), the value resulting from the evaluation on all fitness cases (in other words, it contains the semantics of that individual).  Hence, having a training set with $k$ training instances and $P$ (respectively $P_{\text{mut}}$) containing $n$ (respectively $m$) individuals, results in a table $T_{\text{vp}}$ (respectively $T_{\text{vm}}$) with $k$ rows and $n$ (respectively $m$) columns.

    \item For every generation $p > 1$, a new empty table $T'_{\text{vp}}$ is created and, whenever a new individual $T$ must be generated by crossover between selected parents $T_1$ and $T_2$, the following actions are performed:

    \begin{itemize}

        \item $T$ is represented by a triple $T = \langle \alpha(T_1),\alpha(T_2),\alpha(R)\rangle$, which is stored in an apposite structure (this structure is called ${\mathcal M}$ from now on), where $R$ is one of the random trees in $P_{\text{mut}}$ and, for any tree $\tau$, $\alpha(\tau)$ is a {\it reference} (or memory pointer) to $\tau$.

        \item The first line that is still empty in $T'_{\text{vp}}$ is successively filled with the values of the semantics of $T$, which can be easily obtained by calculating $(T_1 \cdot R) + ((1-R)\cdot T_2)$ for each fitness case, according to the definition of geometric semantic crossover.

    \end{itemize}

    \item Analogously, whenever at generation $p$ a new individual $T$ has to be obtained by applying mutation to an individual $T_1$, the following actions are performed:

    \begin{itemize}

        \item $T$ is represented by a triple $T = \langle \alpha(T_1),\alpha(R_1),\alpha(R_2)\rangle$ (this triple is also stored in ${\mathcal M}$), where $R_1$ and $R_2$ are two among the random trees in $P_{\text{mut}}$.

        \item The first line that is still empty in $T'_{\text{vp}}$ is this time filled with the values of the semantics of $T$ which can be easily obtained by calculating $T_1+ms\cdot(R_1 - R_2)$ for each fitness case, according to the definition of geometric semantic mutation.

    \end{itemize}

    \item When generation $p$ is completed, table $T'_{\text{vp}}$ is copied into $T_{\text{vp}}$ and erased.

    \item The process is iterated for the prefixed number of generations.

\end{itemize}

In synthesis, this algorithm is based on the idea that, when semantic operators are used, an individual can be fully described by its semantics (which makes the syntactic component much less important than in standard GP), a concept discussed in depth in~\cite{PaperMoraglio}.  In order to implement this idea, at every generation we update table $T_{vp}$ by using the values contained in it and the ones in $T_{vm}$, without explicitly building the syntactic structures of the individuals, but incorporating all the information to do it in a second time.

We point out that:

\begin{enumerate}

    \item This process of updating table $T_{vp}$ can be performed efficiently and no evaluation of the whole tree is needed anymore. As a consequence, in this implementation the fitness calculation is rather efficient. Indeed the fitness evaluation process requires, for each individual and except for the first generation, a constant time, which is independent from the size of the individual itself.  On the other hand, at the initial generation, the fitness evaluation requires a time that is dependent on the size of the individuals, as it is usual in GP.

    \item Conceptually, population $P$ evolves during a GP run, while $P_{\text{mut}}$ does not change until the end of each run.  This is implemented by continuously updating table $T_{\text{vp}}$, while $T_{\text{vm}}$ stays unchanged during a run.

    \item The structure ${\mathcal M}$ (that contains, for each individual, the triplet of references to the ancestors) increases in size during a GP run.  However, given that this structure contains only pointers, we can manage it for several thousands of generations in a very efficient way.

    \item The fact that all the random trees used by crossover and mutation are generated in one step before the beginning of the evolutionary process (instead of generating them at the moment they are needed) does not change the expected behaviour of the algorithm (there is no reason to imagine that a random tree should have different properties if generated in two different instants).  Nevertheless, they still can be generated and stored in $T_{\text{vm}}$ at the moment they are needed, instead of doing it all at once in the beginning, with no significant modification in the behavior of the algorithm.

    \item Generating all the random trees that will be needed in one step before starting the evolutionary process and storing them in $P_{\text{mut}}$ is a procedure that can efficiently be managed from a computational viewpoint.

    \item Tables $T_{\text{vp}}$ and $T_{\text{vm}}$ contain the values of the evaluation of the individuals on the fitness cases (i.e. the semantics of the individuals), not their fitness values.  This information (and not the fitness) is the one that is needed to reconstruct the semantics of the individuals in the subsequent generations and iterate the process.  It is nevertheless easy to calculate the fitness using the semantics and knowing the corresponding target values.

\end{enumerate}

The final part of the algorithm has to be performed after the end of the last generation, in order to reconstruct the individuals.  For doing that, we need to ``unwind'' our compact representation and make the syntax of the individuals explicit.  In this way, we will still have the large trees that characterize the standard implementation of geometric semantic operators. However, all the evolutionary process can be performed efficiently and, if we are interested only in the best individual found by GP (which is the typical situation, where the best individual is interpreted as the model explaining data), we can perform the simplification of the expression on only one tree, instead of every tree in the population at each generation as proposed in~\cite{PaperMoraglio}.  Furthermore, the simplification is not performed during the evolution, but it can be done offline in a second step.

Excluding the time needed to simplify the best individual, the proposed implementation allowed us to evolve populations for thousands of generations with a speed up to at least $20$ times higher than standard GP.

In the continuation of this section, we show a simple example that should clarify the functioning of the proposed algorithm.

\subsection{Example}
\label{myex}

Let us consider the simple initial population $P$ shown in Table~\ref{t:P} and the simple pool of random trees $P_{\text{mut}}$ shown in Table~\ref{t:R} (usually $P_{\text{mut}}$ contains a number of individuals much larger than $P$; in this example we consider both $P$ and $P_{\text{mut}}$ containing five individuals for simplicity).
% ----------------------------------------------------------------------------------- -----------------------------------------------------------------------------------
\begin{table}[!ht]
    \center
    \begin{tabular}{|r|l|}
        \hline
        Id &         Individual     \\
        \hline
        $T_1$ &       $x_1 + x_2 \cdot x_3$ \\
        \hline
        $T_2$ &        $x_3 - x_2 \cdot x_4$ \\
        \hline
        $T_3$ &        $x_3 + x_4 - 2 \cdot x_1$ \\
        \hline
        $T_4$ &        $x_3 \cdot x_1$ \\
        \hline
        $T_5$ &        $x_1 - x_3$ \\
        \hline
    \end{tabular}
    \caption{The simple initial population $P$ used in the example of Section~\ref{myex}
      The leftmost column reports the Ids of the individuals. These Ids will be used
      in the text for simplicity.}
    \label{t:P}
\end{table}
% ----------------------------------------------------------------------------------- -----------------------------------------------------------------------------------
\begin{table}[!ht]
    \center
    \begin{tabular}{|r|l|}
        \hline
        Id &         Individual  \\
        \hline
        $R_1$ &       $x_1 + x_2 - 2 \cdot x_4$ \\
        \hline
        $R_2$ &        $x_2 - x_1$ \\
        \hline
        $R_3$ &        $x_1 + x_4 - 3 \cdot x_3$ \\
        \hline
        $R_4$ &        $x_2 - x_3 - x_4$ \\
        \hline
        $R_5$ &        $ 2\cdot x_1$ \\
        \hline
    \end{tabular}
    \caption{The individuals in the random pool $P_{mut}$
      used in the example of Section~\ref{myex}.
      The leftmost column reports the Ids of the individuals. These Ids will be used
      in the text for simplicity.}
    \label{t:R}
\end{table}
% ----------------------------------------------------------------------------------- -----------------------------------------------------------------------------------
Besides the representation of the individuals in infix notation, these tables also contain an Id for each individual ($T_1$, $T_2$, $T_3$, $T_4$ and $T_5$ for the individuals in $P$ and $R_1$, $R_2$, $R_3$, $R_4$ and $R_5$ for the individuals in $P_{mut}$). For simplicity, these Ids will be used from now on to address the different individuals, and individuals that will be created in the subsequent generations will be indicated using letter $T$ follwed by progressive numbers (for example, the five individuals in the population at the second generation will be called $T_6$, $T_7$, $T_8$, $T_9$ and $T_{10}$).

We now describe all the operations involved in the creation of the new population at the next generation, which we indicate as population $P'$ from now on.  Let us assume that the (non-deterministic) selection process imposes that $T_6$ is generated by crossover between $T_4$ and $T_5$.  Analogously, let us assume that $T_7$ is generated by crossover between $T_1$ and $T_4$, $T_8$ is generated by crossover between $T_1$ and $T_5$, $T_9$ is generated by crossover between $T_3$ and $T_4$ and $T_{10}$ is generated by crossover between $T_3$ and $T_5$.  Furthermore, let us assume that to perform these five crossovers, individuals $R_2$, $R_1$, $R_4$, $R_5$ and $R_3$ of $P_{mut}$ have to be used, respectively.

In our implementation, the individuals in $P'$ are simply represented by the set of entries reported in Table~\ref{t:P'}, and stored in structure ${\mathcal M}$.
% ----------------------------------------------------------------------------------- -----------------------------------------------------------------------------------
\begin{table}[!ht]
    \center
    \begin{tabular}{|c|c|c|}
        \hline
        Id & Operator & Entry     \\
        \hline
        $T_6$ & crossover & $\langle T_4,T_5,R_2\rangle$ \\
        \hline
        $T_7$ & crossover & $\langle T_1,T_4,R_1\rangle$ \\
        \hline
        $T_8$ & crossover & $\langle T_1,T_5,R_4\rangle$ \\
        \hline
        $T_9$ & crossover & $\langle T_3,T_4,R_5\rangle$ \\
        \hline
        $T_{10}$ & crossover & $\langle T_3,T_5,R_3\rangle$ \\
        \hline
    \end{tabular}
    \caption{How the individuals in the subsequent generations are stored
      in memory for the example
      of Section~\ref{myex} (this structure is called ${\mathcal M}$
      in the text). The leftmost column reports the Ids of the individuals.
      These Ids will be used
      in the text for simplicity.
      The central column reports the operation that has been used to generate
      the individual (it can be either "crossover" or "mutation". In this example,
      we use only crossover for simplicity). The rightmost column contains references
      to the ancestors used to generate the individual.}
    \label{t:P'}
\end{table}
% ----------------------------------------------------------------------------------- -----------------------------------------------------------------------------------
In synthesis this table contains, for each new individual, a {\it reference} to the ancestors that have been used to generated it and the name of the operator used to generate it (either "crossover" or "mutation").

The only structures that we have to keep in memory during the GP run, besides the ones depicted in Tables~\ref{t:P},~\ref{t:R} and~\ref{t:P'}, are the two tables $T_{\text{vp}}$ and $T_{\text{vm}}$ that contain, at each generation, the values of the evaluation of the individuals in the current population and in $P_{mut}$ for each fitness case.  The size of the structure ${\mathcal M}$ reported in Table~\ref{t:P'} grows during the GP run (in this example, five new entries are added to this table at each new generation, corresponding to the five new individuals in the population); however, it is very compact, because it only contains references, and thus we can manage it for several thousands of generations.

Let us assume that now we want to reconstruct the genotype of one of the individuals in $P'$ (this typically happens only once, at the end of the run, for the best individual in the population). For instance, let us assume that the want to reconstruct $T_8$.  Tables~\ref{t:P},~\ref{t:R} and~\ref{t:P'} provide us with all the information we need to be able to do that. In particular, from Table~\ref{t:P'} we learn that $T_8$ is obtained by crossover between $T_1$ and $T_5$, using random tree $R_4$.  Thus, from the definition of geometric semantic crossover, we know that it will have the following structure: $(T_1 \cdot R_4) + ((1-R_4)\cdot T_5)$.  Table~\ref{t:P}, that contains the syntactic structure of $T_1$ and $T_5$, and Table~\ref{t:R}, that contains the syntactic structure of $R_4$, finally provide us with all the information we need to completely reconstruct the syntactic structure of $T_8$, which is:
$$
((x_1 + x_2 \cdot x_3) \cdot (x_2 - x_3 - x_4)) + ((1-(x_2 - x_3 - x_4))\cdot (x_1 - x_3))
$$
For simplicity, we have omitted mutation in this example and we have generated all the individuals in the new population using only crossover.  Mutation works in a similar way, with the only differences that the central column in Table~\ref{t:P'} contains the label "mutation" (and this information is useful because it tells us that, this time, we have to use the definition of geometric semantic mutation in order to reconstruct the individual) and the triplet associated to the newly generated individual this time contains one reference to an individual in $P$ and two references to two individuals in~$P_{mut}$.

\section{Empirical Study}
\label{sec:results}

\subsection{The Application}

The implementation provided so far makes the geometric semantic operators efficiently usable also on complex real-life applications.  For this reason, for the first time, we are now able to validate those operators on one of those applications.  We choose a real life problem in the field of pharmacokinetic.

As stated in~\cite{springerlink:10.1007/s10710-007-9040-z}, the availability of reliable pharmacokinetics prediction tools would permit to reduce the risk of late stage research failures in drug discovery and will enable to decrease the number of experiments and cavies used in pharmacological research, by optimizing the screening assays. Furthermore, predictive pharmacokinetic models would be of critical relevance for preventing Adverse Drug Reactions (ADRs), like those involved in the Lipobay-Baycol (cerivastatin) toxicity~\cite{tuffs2001}.  The potential of predictive modeling in terms of ADRs prediction is an hot research topic in medicine.  Human oral bioavailability (indicated with \%F from now on) is the parameter that measures the percentage of the initial orally submitted drug dose that effectively reaches the systemic blood circulation after the passage from the liver. This parameter is particularly relevant, because the oral assumption is usually the preferred way for supplying drugs to patients and because it is a representative measure of the quantity of active principle that can actuate its therapeutic effect. Being able to reliably predict the \%F value for a potential new drug is outstandingly important, given that the majority of failures in compounds development from the early nineties to nowadays are due to a wrong prediction of this pharmacokinetic parameter during the drug discovery process~\cite{citeulike:664042,Kennedy1997436}.

We have obtained a set of molecular structures and the corresponding \%F values using the same data as in~\cite{QSAR}, using a public database of Food and Drug Administration (FDA) approved drugs and drug-like compounds~\cite{Wishart06drugbank:a}. The data has been gathered in a matrix composed by $359$ rows and $242$ columns. Each row (instance) is a vector of molecular descriptor values identifying a candidate new drug; each column (feature) represents a molecular descriptor, except the last one, that contains the known values of~\%F.  This dataset can be downloaded from the web page: {\it http://personal.disco.unimib.it/vanneschi/bioavailability.txt}.

In our experiments, training and test sets have been obtained by randomly splitting the dataset: at each GP run, $70\%$ of the molecules have been randomly selected with uniform probability and inserted into the training set, while the remaining $30\%$ form the test set.

\subsection{Experimental Settings}
We tested the proposed implementation of GP with geometric semantic operators (GS-GP from now on) against a standard GP system (STD-GP). A total of $30$ runs were performed with each technique considering different randomly generated partitions of the dataset into training and test set at each run. All the runs used populations of $100$ individuals and the evolution stopped after $2000$ generations. Trees initialization was performed with the Ramped Half-and-Half method~\cite{poli08:fieldguide} with a maximum initial depth equal to $6$. The function set contained the four binary arithmetic operators $+$, $-$, $*$, and $/$ protected as in~\cite{poli08:fieldguide}. Fitness was calculated as the root mean squared error between outputs and targets (thus the lower the fitness, the better the individual). The terminal set contained $241$ variables, each one corresponding to a different feature in the dataset. To create new individuals, STD-GP used standard (subtree swapping) crossover~\cite{poli08:fieldguide} and (subtree) mutation~\cite{poli08:fieldguide} with probabilities equal to $0.9$ and $0.1$ respectively. For GS-GP, crossover rate is $0.9$, while mutation rate is $0.5$. The motivation for this different mutation rate for the two GP systems is that a preliminary experimental study has been performed (independently for the two systems) for finding the parameter setting able to return the best results.  Only the parameter settings that returned the best results for the two systems are presented here.  Survival from one generation to the other was always guaranteed to the best individual of the population (elitism). No maximum tree depth limit has been imposed during the evolution.

In the next section, experimental results are reported using curves of the root mean square error on the training and test set.  In particular, at each generation the best individual in the population (i.e. the one that has the smaller training error) has been chosen and the value of its error on the training and test has been stored. The reported curves finally contain the median of all these values collected at each generation.  The median was preferred over the mean in the reported plots because of its higher robustness to outliers.  The root mean square error on the training and test set, calculated as described above, will be in some cases informally indicated as training and test fitness, or training and test error, in the next section for simplicity.

\subsection{Experimental Results}

\begin{figure*}[!ht]
    \centering \subfigure[]{\includegraphics[width=0.4\textwidth]{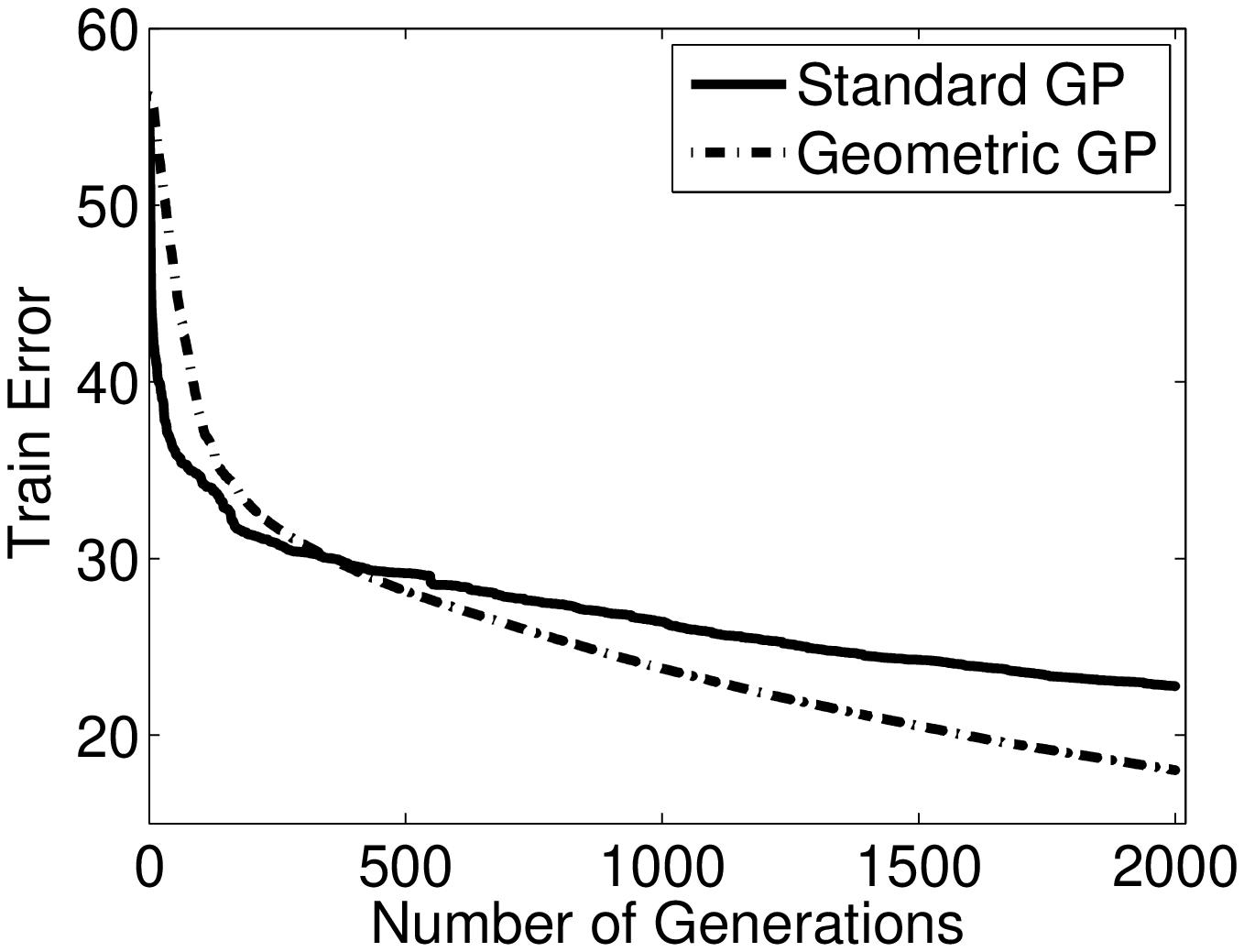}} \subfigure[]{\includegraphics[width=0.4\textwidth]{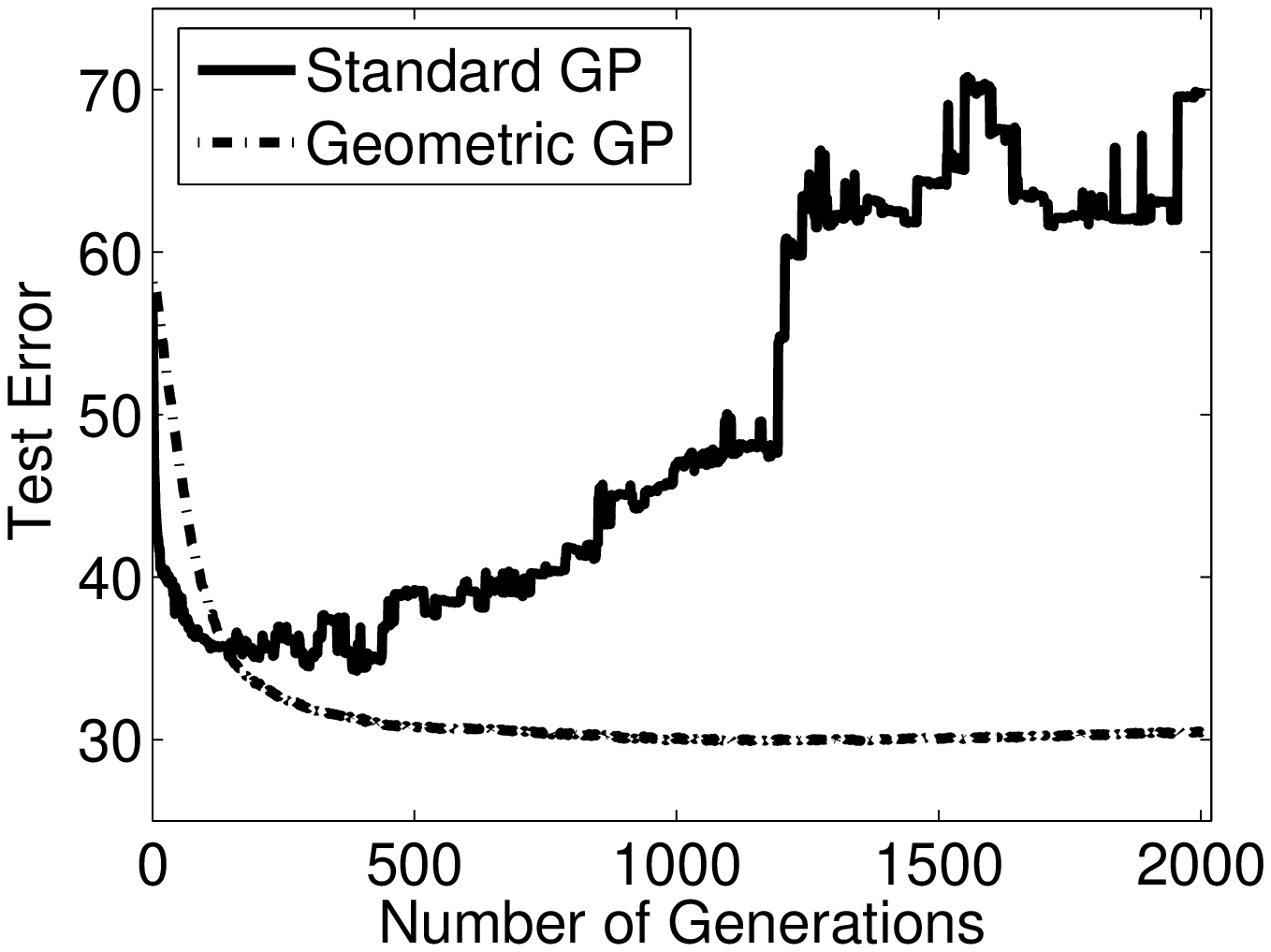}}
    \caption{Median of train and test error for the consided techniques at each generation calculated over $30$ independent runs.}
    \label{fig:error}
\end{figure*}

Figure~\ref{fig:error} reports training and test error for STD-GP and GS-GP and clearly shows that GS-GP outperforms STD-GP on both training and test sets. In particular, GS-GP has a suitable behaviour: the curve of the error on the test set is quite "regular" and steadily decreasing during the whole evolutionary process. This behaviour on the test set gives us a hint of the fact that, contrarily to STD-GP, GS-GP does not overfit training data for the considered application.

\begin{figure*}[!ht]
    \centering \subfigure[]{\includegraphics[width=0.4\textwidth]{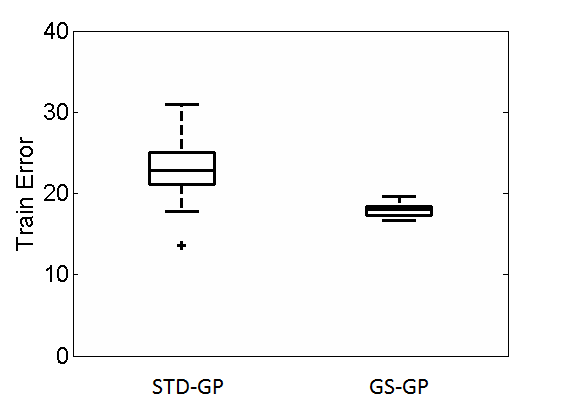}} \subfigure[]{\includegraphics[width=0.4\textwidth]{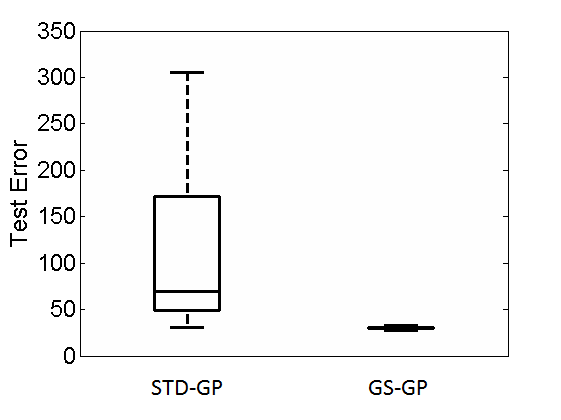}}
    \caption{Train and test error of the best individual produced in each of the $30$ runs at the last performed generation.}
    \label{fig:boxerror}

\end{figure*}

Figure~\ref{fig:boxerror} reports a statistical study of the test fitness of the best individual, both for GS-GP and STD-GP, for each of the $30$ performed runs. Denoting by \emph{IQR} the interquartile range, the ends of the whiskers represent the lowest datum still within $1.5\cdot$ \emph{IQR} of the lower quartile, and the highest datum still within $1.5\cdot$ \emph{IQR} of the upper quartile. As it is possible to see, GS-GP produces solutions with a lower standard deviation with respect to the ones produced by STD-GP. To analyze the statistical significance of these results, a set of tests has been performed on the median errors. As a first step, the Kolmogorov-Smirnov test has shown that the data are not normally distributed and hence a rank-based statistic has been used. Successively, the Wilcoxon rank-sum test for pairwise data comparison has been used under the alternative hypothesis that the samples do not have equal medians. The $p$-values obtained are $6.0\cdot 10^{-11}$ when test fitness of STD-GP is compared to test fitness of GS-GP and $7.1\cdot 10^{-9}$ when training fitness of STD-GP is compared to training fitness of GS-GP. Therefore, when using a significance level $\alpha=0.05$, we can clearly state that GS-GP produces fitness values that are significantly lower (i.e., better) than the STD-GP both on training and test data.

Besides comparing GS-GP with standard GP, we are also interested in comparing GS-GP with other well known state of the art Machine Learning methods, just to have an idea of the competitiveness of the results returned by GS-GP.  Previous studies have appeared so far comparing several Machine Learning techniques for the prediction of the bioavailability of potentially new drugs. For instance, in~\cite{springerlink:10.1007/s10710-007-9040-z} the following methods have been tested: linear regression, least square regression, multilayer perceptron, support vector machines regression with first degree polynomial and support vector machines regression with second degree polynomial kernel.  In~\cite{springerlink:10.1007/s10710-007-9040-z} all these methods are used with and without an explicit feature selection, performed on the original data as a preprocessing phase.  In~\cite{springerlink:10.1007/s10710-007-9040-z} the feature selection methods used are principal component based feature selection and correlation based feature selection.  Here, we take up exactly the same perspective, by using all these methods with and without these feature selection strategies on our dataset.  As in~\cite{springerlink:10.1007/s10710-007-9040-z} we used the implementations provided by the Weka public domain software~\cite{weka} and, for each one of the used Machine Learning methods and feature selection strategies, we have used the default parameter setting of Weka.  The results are reported in Table~\ref{t:MLtech}, where we can observe that the best performance was obtained by linear regression with correlation based feature selection, that returned a root mean square error on the test set approximately equal to 27.52.  Given that GS-GP, in the last performed generation, has returned a median test fitness equal to 30.44, and given that the best test fitness over the performed 30 runs was equal to 26.97, we state that GS-GP is able to find better, or at least comparable, results than the best one of the state of the art Machine Learning methods.  We also point out that these results have been obtained by GS-GP without any explicit feature selection (given that GP is in general able to perform an automatic feature selection during the learning phase~\cite{f4,springerlink:10.1007/s10710-007-9040-z}), while the best results of the state of the art methods have been obtained by explicitly selecting features by the correlation based technique.  The explicit use of a preprocessing phase to select features has also been used so far in Evolutionary Computation in general~\cite{f1}, and in GP in particular~\cite{f3}, with excellent results.  This should further improve GS-GP performances, and new experiments including explicit feature selection are part of our current research.
\begin{table*}[!ht]
    \center
    \begin{tabular}{|l|l|}
        \hline
        Method &  Test RMSE \\
        \hline
        (a) No feature selection & \\
        \hline
        Linear regression & 48.1049 \\
        Least square regression & 37.2211 \\
        Multi layer perceptron & 51.28 \\
        SVM regression-first degree polynomial kernel & 34.804 \\
        SVM regression-second degree polynomial kernel & 44.323 \\
        \hline
        (b) Principal component based feature selection (PCFS) & \\
        \hline
        Linear regression & 30.5568 \\
        Least square regression & 40.4503 \\
        Multi layer perceptron & 48.9771 \\
        SVM regression-first degree polynomial kernel & 36.185 \\
        SVM regression-second degree polynomial kernel & 42.3377 \\
        \hline
        (c) Correlation based feature selection (CorrFS) & \\
        \hline
        Linear regression & 27.5212 \\
        Least square regression & 31.7826 \\
        Multi layer perceptron & 32.5782 \\
        SVM regression-first degree polynomial kernel & 28.8875 \\
        SVM regression-second degree polynomial kernel & 29.7152 \\
        \hline
    \end{tabular}
    \caption{Experimental comparison between different non-evolutionary Machine Learning techniques for oral bioavailability predictions.
      Error on the test reported for each technique.}
    \label{t:MLtech}
\end{table*}

%%%%%%%%%%%%%%%%%%%%%%%%%%%%%%%%
% Conclusions and Future Works %
%%%%%%%%%%%%%%%%%%%%%%%%%%%%%%%%
\section{Conclusions and future works}
\label{sec:conclusioni}

New genetic operators, called geometric semantic operators, have been defined so far for genetic programming.  They have the extremely interesting property of inducing a unimodal fitness landscape for any problem consisting in matching input data into known output ones (regression and classifications are instances of this general problem).  This, at least at a theoretical level, should make all the problems of this kind easily solvable by genetic programming.  Nevertheless, as demonstrated in the literature, these new operators, in their current definition, have a strong limitation, that makes them unusable in practice: they produce offspring that are larger than their parents, and this comports an exponential growth in the size of the individuals in the population.

The goal of this paper is to overcome this limitation by proposing a new genetic programming system, in which geometric semantic operators are implemented in a very efficient way.  The proposed implementation basically keeps in memory only the initial (randomly generated) population of programs, plus a set of randomly generated programs that will be used by the operators during the evolution.  Furthermore, the implementation stores and maintains updated some tables containing pointers to those programs.  The size of these tables grows linearly with generations, and thus managing those tables is quite feasible.

Thanks to this compact and efficient implementation, it is possible, for the first time, to employ the framework to solve complex problems, characterized by a large number of features. In particular, in this paper an important real life problem in the field of pharmacokinetic has been considered.

The presented experimental results demonstrate that the new system outperforms standard genetic programming and returns results that are better, or at least comparable to the best state of the art machine learning method for this application.  Besides the fact that the new genetic programming system has excellent results on training data (which was expected, given that the fitness landscape is unimodal), we are positively surprised by its excellent generalization ability on the studied application, testified by the good results we have obtained on test data.

This encourages us to pursue the study of geometric semantic operators on real-life applications.  Furthermore, we also plan to orient our future activity towards more theoretical studies of geometric semantic operators, looking for formal models able to explain the generalization ability of the new genetic programming framework.

\bibliographystyle{IEEEtran}
\bibliography{IEEEabrv,bibliografia}

\end{document}